# A New Characterization of Probabilities in Bayesian Networks

**Lenhart K. Schubert**

Computer Science Dept.
University of Rochester
Rochester, NY 14627-0226

## Abstract

We characterize probabilities in Bayesian networks in terms of algebraic expressions called quasi-probabilities. These are arrived at by casting Bayesian networks as noisy AND-OR-NOT networks, and viewing the subnetworks that lead to a node as arguments for or against a node. Quasi-probabilities are in a sense the "natural" algebra of Bayesian networks: we can easily compute the marginal quasi-probability of any node recursively, in a compact form; and we can obtain the joint quasi-probability of any set of nodes by multiplying their marginals (using an idempotent product operator). Quasi-probabilities are easily manipulated to improve the efficiency of probabilistic inference. They also turn out to be representable as square-wave pulse trains, and joint and marginal distributions can be computed by multiplication and complementation of pulse trains.

## 1 Introduction and preliminaries

The work reported here began as an attempt to interpret Bayesian networks (henceforth BNs) as static representations of arguments for or against the propositions denoted by the nodes. The hope was that such a perspective would indicate how BN-style reasoning could be "lifted" to first-order probabilistic reasoning. Of course, there has been a sustained effort to amalgamate BN inference with first-order logic (e.g., Goldman & Charniak 1990, Wellman *et al.* 1992, Poole 1993, Haddawy 1994, Ngo & Haddawy 1996, Poole 1997, Jaeger 1997, Koller 1998, Cussens 1999, Kersting & de Raedt 2000, Pfeffer 2000, Pasula & Russell 2001, Poole 2003), but the proposed methods are typically limited to Horn logic (often function-free, and often range-restricted) at least when reasoning probabilistically, and BN-style inference is implemented by explicit query-driven construction of BNs. What we are seeking instead is a probabilistic logic whose inference methodology more closely resembles that of FOL: facts and rules (in particular causal rules) should be usable one at a time for forward and backward inference. Probabilities or bounds on them would be subject to revision (but convergent), with tacit use of BN-like independence assumptions.

The results obtained so far fall short of that larger goal, but they seem both theoretically interesting and potentially useful. The interpretation of BNs as argument structures leads to a very natural algebraic characterization of propositional probabilities in BNs. We develop the basic properties of these algebraic "quasi-probabilities" here, and also suggest various ways in which they might be exploited for BN inference (e.g., diagnosis, plan projection, and SAT-solving). We will also return briefly to the issue of first-order BN-like inference at the end.

### 1.1 AND-OR-NOT Bayesian networks

We begin with the notion of a *symbolically labelled, noisy AND-OR-NOT Bayesian network* (or an *AND-OR-NOT BN* for short). Except for the roots, the nodes of such a BN are interpreted as noisy AND, OR, or NOT gates. A *noisy AND node* has 2 or more parents, and we draw an arc across the incident links to distinguish AND nodes from OR nodes. An AND node can be true only if *all* its parents are true; and the conditional probability that an AND node is true, given that its parents are true, is signified by a label $p$ placed on the combined links from the parents to the node. This label is either 1 or a distinct *elementary probability* symbol. A *noisy OR node* is defined (as usual) as being independently influenced by its parents. Each parent, if true, has some probability $p$ of "causing" the OR node to be true, when all the other parents are false; the label $p$, which may again be 1 or a distinct elementary probability symbol, is placed on the link from the parent. When multiple parents are true, say those with link labels $p_1, p_2, ..., p_k$, then the (conditional) probability of truth of the OR node is

$$1 - (1 - p_1)(1 - p_2)...(1 - p_k).$$

We make the *accountability* assumption (Pearl 1988) that an OR node is false whenever all its parents are false.

We take all root nodes of an AND-OR-NOT BN as having probability 1, for convenience. (Obviously, any root $R$ with nonunit probability $r$ can be modelled by adding a unit-probability root $R'$ and a single link labelled $r$ from $R'$ to $R$.) Finally, a *noisy NOT node* is a single-parent node that is false whenever its parent is true, and that has some probability $p$ of being true when its parent is false. The link from the parent is drawn as an "inhibitory" link, with a small circle in place of the arrowhead, and the label may again be 1 or a distinct elementary probability symbol. We also use a compact notation that allows inhibitory links to AND nodes and OR nodes. An AND node is true with probability $p$ (the joint link label) if its inhibitory links come from false parents and its ordinary links come from true parents (and is false otherwise). An inhibitory link labelled $p$ from a false parent to an OR node affects that node exactly like an ordinary link labelled $p$ from a true parent. It is not hard to see that inhibitory links to AND nodes and OR nodes can be eliminated by introducing extra NOT nodes.

The quasi-probability characterization we develop below for AND-OR-NOT BNs actually applies to all boolean-valued BNs, since the latter are easily converted to the former. This is illustrated for the case of a 2-parent node in Fig. 1.

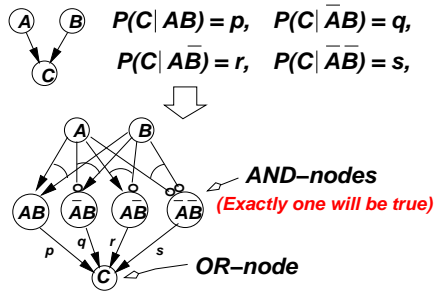

**Figure 1.** Mapping an arbitrary BN node to an AND–OR–NOT network.

Note that we are introducing distinct atomic symbols for each of the probability parameters (except those = 0 or 1). The AND nodes have probability 1. The transformation is easily generalized to nodes with indegree $n$. Essentially what we are doing is to convert a conditional probability table of size $2^n$ into $2^n$ explicit nodes. Of course, the elementary probability symbols introduced by the transformation would ultimately be instantiated to the given numerical values when we use the BN for inference of numerical probabilities.

In preparation for the connection we will make between conventional representations of BN probabilities and quasi-probabilities, we note the following recursion equations for AND nodes and OR nodes (together with an arbitrary set of additional nodes) in an arbitrary BN. Note that these characterizations of the probability of truth make no reference to falsity of any nodes.

**Proposition 1** (*P*-recursion for AND). Let $\mathcal{E}$ be a set of nodes in a BN and let $C$ be a binary AND node which is not in $\mathcal{E}$ and has no descendants in $\mathcal{E}$. Let $C$'s parents be $A_1, ..., A_n$, whose links to $C$ are jointly labelled $r$. Then

$$P(\mathcal{E}C) = rP(\mathcal{E}A_1...A_n).$$

Here $P(\mathcal{E}C)$ is the probability that all nodes in $\mathcal{E}$ as well as the node $C$ are true, and analogously for $P(\mathcal{E}A_1...A_n)$.

**Proposition 2** (*P*-recursion for OR). Let $\mathcal{E}$ be a set of nodes in a BN and let $D$ be an OR node which is not in $\mathcal{E}$ and has no descendants in $\mathcal{E}$. Let the parents of $D$ be $A_1, ..., A_n$ ($n \geq 1$), whose links to $D$ are respectively labelled $r_1, ..., r_n$. Then

$$P(\mathcal{E}D) = \sum_{b \in \mathcal{B}_n} odd(b)\Big(\prod_{b_i = 1} r_i\Big) P(\mathcal{E}\{A_i\}_{b_i=1}).$$

where *odd* is a sign-function on bit-vectors $b \in \mathcal{B}_n = \{b_1...b_n | b_1, ..., b_n \in \{0, 1\}\} \setminus \{0...0\}$, such that for any term $\tau$, $odd(b)\tau = \tau$ if $\sum_{i=1}^{n} b_i$ is odd, and $= -\tau$ otherwise.

The latter (less obvious) proposition can be proved by induction on $n$, using a decomposition of a $(k+1)$-ary OR node into a $k$-ary and a binary OR node in the induction step.

## 1.2 From arguments to quasi-probabilities

At this point we take note of a curious "coincidence" concerning probabilities in AND-OR-NOT BNs – one that in fact led to the new characterization of probabilities in such nets.

Consider the examples shown in Fig. 2. In both networks, loops have been drawn around those portions that can be viewed as (probabilistic) *arguments* for the truth of the bottom node. In the BN on the left, we can argue that $A$ is true, hence $B$ may be rendered true, hence both $C$ and $D$ may be rendered true, and since $E$ is an AND node, this may render $E$ true. The total probability of the argument is simply the product of all the elementary probabilities it involves, i.e.,

*pqrs.*

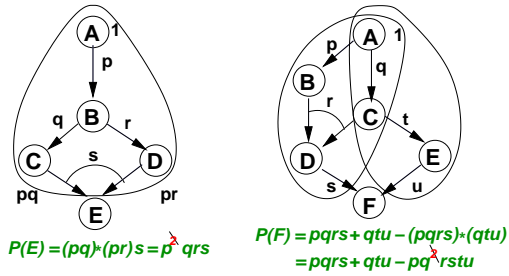

*Figure 2.* "Arguments" for the truth of nodes

However, if we take a more incremental view (arguing in step-by-step logical fashion), we find that $C$ has probability $pq$ and $D$ has probability $pr$. If we then try to take the final step by treating $C$ and $D$ as independent, we naturally get the wrong result, $p^2qrs$. In fact, the interdependence of $C$ and $D$ is evidenced by the occurrence of $p$ in both their probabilities – a consequence of their common ancestry. But we can "correct" for the error by collapsing higher powers of $p$ to $p$. In other words, we treat the multiplication operator marked in the figure as "*" as *idempotent* (it is analogous to logical $\wedge$, e.g., $p \wedge q \wedge p \wedge r \equiv p \wedge q \wedge r$ ).

In the BN on the right, there are two possible arguments for the truth of $F$, since $F$ is a binary OR node. In this case, we have indicated what would happen if we treated the two arguments as independent. Since their individual probabilities are clearly $pqrs$ and $qtu$ respectively, we would arrive at the indicated noisy-OR-like combination $pqrs + qtu - (pqrs) * (qtu)$, and again, by treating "*" as idempotent rather than ordinary product, we can obtain the correct result in this way. Algebraic probabilities like those just illustrated, with an idempotent product operator, will shortly be formalized as quasi-probabilities.

In fact it turns out that the probability of truth of any set of nodes $\mathcal{E}$ of an AND-OR BN can be expressed as

$$P(\mathcal{E}) \simeq 1 - \prod_{i=1}^{N}{}^{*}(1 - \rho_i),$$

where $\rho_1, ..., \rho_N$ are the probabilities of the $N$ distinct arguments for the truth of $\mathcal{E}$, expressed as products of elementary probabilities. Crucially, the iterated product on the right is based on the "*" operator, as indicated by the superscript. The notion of an argument for the truth of a node can also be extended to AND-OR-NOT BNs, by simultaneously defining the notion of an argument for the falsity of a node, but we set this aside.[1]

---

[1] The latter notion is not so simple, since for example a single-parent OR node can be false either because the parent is false, or because the influence of the parent failed to have an effect, though the parent is true.

Although we make no explicit use of argument subnetworks in what follows, the informal observations we have made about combining probabilities of nodes or node sets as if they were independent, while using an idempotent product operator, provides the intuitive basis for our formal development.

## 2  Quasi-probabilities (QPs) and BN probabilities

We define quasi-probabilities below. They are much like ordinary algebraic expressions based on $+$, unary and binary $-$, and $\cdot$ (arithmetic product), except that "*" takes the place of "$\cdot$". We will term "*" the *weak product* operator since it forms no powers of elementary probabilities. The prefix "*quasi*" is intended to suggest that quasi-probabilities cannot be directly evaluated numerically, as long as some identical elementary probabilities occur on both sides of a weak product operator.

*Definition (abbreviated).* The class of *quasi-probabilities* (QPs) based on a set $Q = \{q_1, q_2, ...\}$ of elementary probabilities (which we can always identify for our purposes with the labels of a particular BN) consists of the following expressions:

(a) 0, 1, or $p_1p_2...p_k$, where $k \geq 1$ and the $p_i$ are *distinct* elementary probabilities;

(b) any expression $(1 - \tau)$ where $\tau$ is a QP;

(c) any expression $(\tau_1 * \tau_2)$, where $\tau_1, \tau_2$ are QPs; (*weak product*);

(d) any expression $\tau[\sigma]$ obtained from a QP $\tau[\rho]$, where $[\rho]$ indicates an occurrence of a subexpression $\rho$ somewhere in $\tau$ and $\tau[\sigma]$ is the result of replacing that occurrence by $\sigma$, and $\rho$ and $\sigma$ are related in one of 14 ways:

   (i) (product reduction) if $\rho$ is $(p_1...p_k * q_1...q_\ell)$, where the $p_i$ and $q_j$ are elementary probabilities, then $\sigma = r_1...r_m$, where all $r_i$ ($1 \leq i \leq m$) are distinct and $\{r_1, ..., r_m\} = \{p_1, ..., p_k\} \cup \{q_1, ..., q_\ell\}$;

   (ii) ($+-$ introduction) if $\rho$ is of form $(\rho_1 - \rho_2)$ then $\sigma = (\rho_1 + (-\rho_2))$;

(iii-xiv) the following informal enumeration of the remaining allowable substitutions (used because of space limitations) should allow their formal reconstruction: commutativity and associativity of $*$ and $+$ (e.g., if $\rho$ is of form $(\rho_1 * \rho_2)$ then $\sigma = (\rho_2 * \rho_1)$; $*$-distribution over $+$; extraction of unary $-$ out of a product, its distribution over $+$, and $--$ elimination; and simplification of $(\rho_1 - \rho_1)$, $(0 * \rho_1)$, $(0 + \rho_1)$ and $(1 * \rho_1)$ as expected (e.g., if $\rho$ is of form $(\rho_1 - \rho_1)$ then $\sigma = 0$).

*Definition.* Two QPs $\sigma$, $\tau$ are *equivalent*, written as $\sigma \simeq \tau$, if they can be reduced to identical expressions using the reduction operations in (d), along with permutation of elementary probabilities occurring in subexpressions that are simple QPs.[2]

Note that by definition every QP is equivalent to one involving only $1-$ (subtraction from 1) and $*$. However, this form (which might be called *negation-conjunction form*, because of the close correspondence of $1-$ to negation and $*$ to conjunction) is insufficiently flexible for making the link to numerical BN probabilities. For example, the QP $[1 - p(1 - q)] * (1 - qr)$ can be numerically evaluated in the equivalent form $1 - qr - p(1 - q)$, where no atom occurs in both factors of a product, but no such "decomposed" form exists that involves only $1-$ and $*$.

We streamline our notation in the usual way by dropping brackets where no ambiguity can result, under the operator precedence ordering $* \succ - \succ +$, and the assumption that extended sums and products are to be read left-associatively. As already indicated, we use $\prod^*$ for the iterated application of the weak product operator $*$ (and we take $\prod^* \emptyset = 0$).

Some noteworthy and useful properties of QPs are provided by the following three lemmas.

**Lemma 1** (expansion lemma). If $\rho_1, ..., \rho_n$ are QPs, then

$$1 - \prod_{i=1}^{n}{}^* (1 - \rho_i) \simeq \sum_{b \in \mathcal{B}_n} odd(b) \prod_{b_i=1}^{n}{}^* \rho_i,$$

where *odd* is defined as in Proposition 2. This can be proved by induction on $n$, and does not depend on the "product reduction" (idempotency) properties of $*$, only on the properties it shares with ordinary multiplication.

QPs are idempotent by definition at the level of elementary probabilities (see defining property (d)(i)), but actually turn out to be uniformly idempotent:

**Lemma 2** (idempotency). For any QP $\tau$, $\tau * \tau \simeq \tau$.

This property of QPs (which can be proved by induction on the complexity of $\tau$) is the key to their usefulness in characterizing probabilities in BNs.

Lemma 3 allows radical simplification of certain kinds of products.

**Lemma 3** (decoupling). Where $\rho$ and $\rho_1, ..., \rho_n$ are

QPs,

$$\prod_{i=1}^{n}{}^* (1 - \rho * \rho_i) \simeq 1 - \rho * \Big[1 - \prod_{i=1}^{n}{}^* (1 - \rho_i)\Big].$$

The proof is by induction on $n$ and use of idempotency.

We now define the QP of a set of nodes in an AND-OR-NOT BN, with the goal of showing that this QP is *correct*, i.e., equivalent to the probability entailed by the definition of such BNs. Essentially the definition is by analogy with the AND-recursion and OR-recursion equations of Propositions 1 and 2 – and indeed these are the key (along with idempotency and other properties of QPs) to establishing correctness.

*Definition.* Given an AND-OR-NOT BN, the *quasi-probability of truth*, $P^*(\mathcal{E})$, of a nonempty set of nodes $\mathcal{E}$ of the BN is defined as follows:

- If $\mathcal{E}$ is a set of roots, then $P^*(\mathcal{E}) = 1$.

- If $\mathcal{E} = \{C\}$, where $C$ is a node with parents $A_1, ..., A_n$, then $P^*(C)$ is determined as follows, depending on the node type:

  AND: $P^*(C) = p * P^*(A_1, ..., A_n)$, where $p$ is the joint label of the links into $C$;

  OR: $P^*(C) = 1 - \prod_{i=1}^{n}{}^* \Big(1 - p_i * P^*(A_i)\Big)$, where $p_i$ is the label of the link from $A_i$ to $C$, for $1 \le i \le n$;

  NOT: $P^*(C) = p * (1 - P^*(A_1))$, where $p$ is the label of the (inhibitory) link from $A_1$ to $C$.

- Otherwise, for $\mathcal{E} = \{C_1, ..., C_n\}$ $(n > 1)$, $P^*(\mathcal{E}) = \prod_{i=1}^{n}{}^* P(C_i)$.

$P^*(\mathcal{E})$ is well-defined since the definition determines it uniquely (in terms of elementary probabilities in the BN) apart from the ordering of weak product operations, but this ordering is immaterial in view of the commutativity of $*$. The following is our central result.

**Theorem 1.** For any set of nodes $\mathcal{E}$ of an AND-OR-NOT BN (with $\mathcal{E} \ne \emptyset$), $P^*(\mathcal{E}) \simeq P(\mathcal{E})$.

**Proof sketch.** The proof uses induction on the maximal topological index among nodes of $\mathcal{E}$, in a topological sort of the network, and separately considers the cases of a root node (which is the basis), AND node, OR node, and NOT node. We will show the induction induction step for AND and some glimpses of the induction step for OR.

Let $C$ be an AND node with parents $A_1, ..., A_n$, whose links to $C$ are jointly labelled $r$. Then

---

$$P^*(\mathcal{E}C) \simeq P^*(\mathcal{E}) * P^*(C) \simeq P^*(\mathcal{E}) * r * P^*(A_1...A_n)$$
$$\simeq r * P^*(\mathcal{E}A_1...A_n) \simeq r * P(\mathcal{E}A_1...A_n)$$
$$\simeq r P(\mathcal{E}A_1...A_n) = P(\mathcal{E}C).$$

The first line uses the definition of $P^*$; the second does as well, and also uses the commutativity of "$*$", Lemma 2 (idempotency – required since some of the $A_i$ may occur in $\mathcal{E}$), and the induction assumption; and the third line uses the fact that $r$ does not occur in $P(\mathcal{E}A_1...A_n)$, and Proposition 1 (last step).

Let $C$ be an OR node with parents $A_1, ..., A_n$, whose links to $C$ are respectively labelled $r_1, ..., r_n$. Then some steps of the induction argument are as follows.

$$P^*(\mathcal{E}C) \simeq P^*(\mathcal{E}) * P^*(C) \text{ by definition}$$
$$\simeq P^*(\mathcal{E}) * \left[1 - \prod_{i=1}^{n}{}^* \left(1 - r_i * P^*(A_i)\right)\right] \text{ by def}^n$$
$$\simeq \sum_{b \in \mathcal{B}_n} \left[odd(b) \left(\prod_{b_i=1}{}^* r_i\right) * P^*(\mathcal{E}\{A_i | b_i = 1\})\right]$$
$$\simeq \sum_{b \in \mathcal{B}_n} \left[odd(b) \left(\prod_{b_i=1} r_i\right) P(\mathcal{E}\{A_i | b_i = 1\})\right]$$
$$= P(\mathcal{E}C) \text{ by Proposition 4 } (P\text{-recursion for OR}).$$

The third line is obtained by use of Lemma 1 (expansion), then commutativity of "$*$", then the definition of $P^*$, and then Lemma 2 (idempotency). The fourth line is obtained by the induction assumption, and the fact that the $r_i$ are distinct and do not occur in $P(\mathcal{E}\{A_i | b_i = 1\})$. We omit the induction step for NOT nodes, which is straightforward. □

It is now easy to see as well how we can compute the probability of any combination of truth values of a subset of nodes of an AND-OR-NOT Bayes net, in terms of QPs:

**Corollary 1.** Let $\{A_1, ..., A_n\}$ $(n \geq 1)$ be any subset of nodes of an AND-OR-NOT BN and consider any bit-vector $b \in \mathcal{B}_n$. Then

$$P(\{A_i\}_{b_i=1} \{\overline{A_i}\}_{b_i=0}) \simeq$$
$$\left(\prod_{b_i=1}{}^* P^*(A_i)\right) * \left(\prod_{b_i=0}{}^* (1 - P^*(A_i))\right)$$

This is easily proved by imagining each node $A_i$ with $b_i = 0$ to have an inhibitory link, with label 1, to a fictitious NOT node.

Finally we note that the joint probability of any set of nodes can be obtained simply as the weak product of their marginals (we might call this "quasi-independence" of all nodes):

**Corollary 2.** For any set of nodes $\mathcal{E} = \{C_1, ..., C_n\}$ of an AND-OR-NOT BN,

$$P^*(\mathcal{E}) \simeq P(\mathcal{E}) \simeq \prod_{i=1}^{n}{}^* P(C_i).$$

Thus a conditional probability $P(A_1...A_m | C_1...C_n)$, where the $A_i$ and $C_j$ are nodes (or negations of nodes), is given by the *conditioning formula*

**(C)** $P(A_1...A_m | C_1...C_n) \simeq$
$$\frac{P^*(A_1) * ... * P^*(A_m) * P^*(C_1) * ... * P^*(C_n)}{P^*(C_1) * ... * P^*(C_n)}.$$

It will become apparent from the discussion in the following section at what point the division indicated by the horizontal bar can be carried out.

## 3 Calculating with QPs

Computational properties of the new characterization are largely a matter for further research, so the discussion here is necessarily preliminary. We integrate some comments on related work with this discussion. We briefly consider, in turn, optimization of exact probability calculations in BNs, and a novel computational approach based on square wave pulse trains.

### 3.1 Optimization of exact probability calculations

Deriving numerical values from QPs requires that we first confine occurrences of any term $\tau$ (containing atomic elements other than 0 and 1) to at most one factor of any weak product. For example, while $(1-p)*(1-q)$ can be rewritten as $(1-p)(1-q)$ and directly evaluated, $(1-p)*(1-pq)$ cannot, since $p$ occurs in both factors. Splitting this into $(1-p)-(1-p)*pq$ shows that it is equivalent to $(1-p)$, since $(1-p)*p \simeq 0$.[3]

In general, $*$-elimination from a QP should be performed so as to limit growth of the expression as much as possible. Of course, we have to reckon with exponential growth in the worst case, since BN inference (and even approximation) is NP-hard (Cooper 1990, Dagum & Luby 1993). But, depending on the structure of the network, we can often radically limit the growth of the transformed QP. Note for instance that for an AND-OR-NOT polytree, the marginal QP of any node allows immediate replacement of all $*$-occurrences by ordinary product. (Conditioning complicates matters, but remains polynomial-time.[4])

---

[3] However, note that in the conditioning formula (C) we may be able to divide out certain QPs without full $*$-elimination, viz., any subset of the $P^*(C_j)$ terms sharing no elementary probabilities with other such terms or the $P^*(A_i)$ terms.

[4] Every $P^*(X)$ or $(1 - P^*(X))$ factored into the numerator of (C) will either share no elementary probabilities with

The following are several rules that we can employ in *-elimination. We will say that two QPs (or subexpressions of QPs) are *unrelated* if none of the elementary probabilities that occur in one (other than perhaps 0 or 1) occur in the other.

1. (*Book-keeping*) We can rewrite a weak product of form $\sigma * \tau$, where $\sigma$ and $\tau$ are unrelated, as $\sigma\tau$. Though we could leave the "*" in place, this rule keeps track of the fact that the elementary probabilities in $\sigma$ don't occur elsewhere.

2. (*Resolution*) We can rewrite a weak product of form $\sigma * \tau$ as $\sigma * \tau[1/\sigma]$, where $[1/\sigma]$ indicates the substitution of 1 for all occurrences of $\sigma$ in $\tau$. This useful equivalence is the direct result of idempotency. By writing $(1-\sigma)$ in place of $\sigma$, we also see that $(1-\sigma)*\tau$ can be rewritten as $(1-\sigma)*\tau[0/\sigma]$.

3. (*Decoupling*) We can apply Lemma 3 to rewrite a product of form

$$(1 - \rho * \rho_1) * ... * (1 - \rho * \rho_n) \quad \text{as}$$
$$\{1 - \rho * [1 - (1 - \rho_1) * ... * (1 - \rho_n)]\}.$$

We also have an ordering rule for distributing "*" over sums/differences, but omit it for brevity. We have found these rules to be effective in various examples but have not yet developed them into an algorithm (say, of the "greedy" variety). We would expect that the optimizations that could be obtained by such an algorithm would be similar to those obtainable by combinatorial optimization methods based on network structure (e.g., Lauritzen & Spiegelhalter 1988, Shachter *et al.* 1990, Jensen & Jensen 1994, Li & D'Ambrosio 1994).

As a simple illustration of some of the rules, let us return to the right-hand network in Fig. 2, and compute $P(B|F)$ using (C). Applying the recursive definition of QPs to node $F$, we obtain (with implicit use of rule (1))

$$P^*(F) \simeq [1 - (1 - sP^*(D)) * (1 - uP^*(E))]$$
$$\simeq [1 - (1 - rsP^*(B) * P^*(C)) * (1 - utP^*(C))]$$
$$\simeq [1 - (1 - pqrs) * (1 - qtu)].$$

By rule 3 applied to $q$ in the two factors above,

$$P^*(F) \simeq 1 - \{1 - q[1 - (1 - prs)(1 - tu)]\}$$
$$\simeq q[1 - (1 - prs)(1 - tu)].$$

This provides the denominator in (C). The numerator is then

$$P^*(B) * P^*(F) \simeq p * q[1 - (1 - prs)(1 - tu)]$$
$$\simeq pq[1 - (1 - rs)(1 - tu)],$$

other factors, or where it does, the shared occurrences are again embedded in terms of form $P^*(X)$ or $(1 - P^*(X))$; likewise for the denominator. Thus rules (1) & (2) in this subsection allow immediate simplification.

by rule (2) applied to $p$ and the QP to the right of the "*". Thus (C) yields (dividing out the $q$)

$$P(B|F) = \frac{p[1 - (1 - rs)(1 - tu)]}{[1 - (1 - prs)(1 - tu)]},$$

which may now be evaluated numerically.

To provide a slightly more complex illustration that also serves to introduce an application domain interesting in its own right, we consider *SAT-solving*. A more or less self-explanatory example is shown in Fig. 3. The QP of the formulas in the figure can be written and reduced as follows. (Use of rule (1) is implicit.)

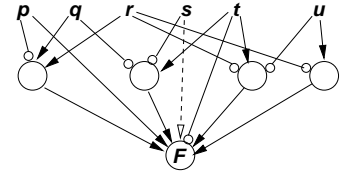

**The bottom node represents the formulas {~pVqVr, p, ~qV~sVt, ~t, ~rVtV~u, ~rVu).**

**If s is added, the formulas become unsatisfiable**

**Figure 3.** Deciding satisfiability

$$P^*(F) \simeq [1 - p(1-q)(1-r)] * p * [1 - qs(1-t)] *$$
$$(1-t) * [1 - r(1-t)u] * [1 - r(1-u)]$$
$$\simeq p(1-t)[1 - (1-q)(1-r)] * (1 - qs) *$$
$$(1 - ru) * [1 - r(1-u)]$$
$$\text{by (2), applied to } p \text{ and } (1-t)$$
$$\simeq p(1-t)[1 - (1-q)(1-r)] * (1 - qs) *$$
$$\{1 - r[1 - (1-u) * u]\}$$
$$\text{by (3) applied to the last 2 factors}$$
$$\simeq p(1-t)[1 - (1-q)(1-r)] * (1 - qs) *$$
$$(1 - r) \text{ since } u * (1-u) = 0 \text{ by (2)}$$
$$\simeq p(1-t)(1-r)[1 - (1-q)] * (1 - qs) \text{ by (2)}$$
$$\simeq p(1-t)(1-r)q * (1 - qs)$$
$$\simeq pq(1-t)(1-r)(1-s) \text{ by (2)}$$

Since the final product contains only isolated occurrences of each of the variables, it is not identically 0, and hence the set of formulas is satisfiable. Further, a satisfying assignment (in fact the only one) is one that assigns truth to $p$ and $q$ and falsity to $r$, $s$, and $t$. If we had added, e.g., $s$ to the set of formulas (as indicated in the figure), the final product would have come to $pq(1-t)(1-r)s*(1-s)$, which is identically 0.

Evidently, what we have here is a (sketchy) decision procedure for SAT, and it will be interesting to investigate how it relates to other procedures such as DPLL (Davis *et al.* 1962) and the DNNF-based method of (Darwiche 2001). (One salient point is that our "reso-lution" rule (2) is related to the 1-literal rule and resolution rule in DPLL, and to "conditioning" in Darwiche's method.) The example also more generally suggests a view of reasoning as manipulating symbolic propositional probabilities, a point to which we will return in the concluding discussion.

Readers familiar with Poole's probabilistic Horn abduction (PHA) (Poole 1993) may suspect a close connection between QPs and sets of *explanations* in PHA, since explanations seem related to argument subnetworks such as those in Fig. 2. However, the connection turns out to be rather indirect in general. In PHA, the parameters $p, q, r, ...$ would be treated as probabilities of choice variables (root nodes), say $P, Q, R, ...$ respectively. Explanations are irredundant, mutually exclusive sets of choices that entail the data (node values) to be explained. In the left-hand network of Fig. 2, the unique explanation of the truth of $E$ (in Poole's notation) would be $[P, Q, R, S]$, with probability $pqrs$. While in this case the correspondence to the (unique) argument for the truth of $E$ is close, in the right-hand network we have six explanations for the truth of $F$: $[P, Q, R, S, T, \overline{U}]$, $[P, Q, R, S, \overline{T}]$, $[\overline{P}, Q, T, U]$, $[P, \overline{R}, Q, T, U]$, $[P, R, \overline{S}, Q, T, U]$, and $[P, Q, R, S, T, U]$. Hence the probability of $E$ is calculated as $pqrst(1 - u) + pqrs(1 - t) + (1 - p)qtu + p(1 - r)qtu + pr(1 - s)qtu + pqrstu$. Naturally, this can be viewed as a QP equivalent to the one obtained by our recursive definition, viz., $1 - (1 - pqrs) * (1 - qtu)$ (if it could not, either the QP calculus or PHA would be incorrect!) But what distinguishes QP representations is their compactness (in general, conversion of the marginal QP of a BN node to a set of explanations is exponentially complex), and the opportunities the QP algebra offers for probability calculation by algebraic manipulation.

Finally, we note that among combinatorial optimization approaches to BN inference, Darwiche's differential approach (Darwiche 2003) seems particularly closely related to one based on QPs. In fact, Darwiche's general polynomial for a BN, for any instantiation of its "indicator" variables fixing the values of some subset of nodes, yields a polynomial for the joint probability of those values.[5] The difference from the QP of those same values is that Darwiche's polynomials are by definition in a fully expanded form, essentially a sum over instantiations of the complete joint p.f. of the BN. But Darwiche builds an arithmetic circuit for his general polynomial, embedding the circuit in a join tree to optimize its computational properties. He then obtains joint probabilities by differentiation and evaluation of the arithmetic circuit. From this perspective, our rules for ∗-elimination listed above can perhaps be viewed as ways to derive an efficient arithmetic circuit for a particular joint probability.

## 3.2 Pulse trains and QPs

One intriguing possibility for computing probabilities in BNs arises from the following simple observations about square-wave pulse trains, where the height of a wave at any time is 0 or 1.

First of all, note that the point-by-point product of a pulse train $\sigma$ with itself is $\sigma$; in other words, such pulse trains are idempotent under point-by-point product. Other analogies to QPs are also immediately apparent; for example, $\sigma \cdot (\mathbf{1} - \sigma) = \mathbf{0}$, (where $\mathbf{1}$ and $\mathbf{0}$ denote uniform "waves" of height 1 and 0 respectively, and "." denotes the point-by-point product). Furthermore, the point-by-point product of two "uncorrelated" square-wave pulse trains has an area (per unit distance along the horizontal axis) that is the product of the areas under the two pulse trains. Given any pulse train with fixed pulse spacing $d$ and pulse width $w < d$, we can ensure that it is not correlated with other pulse trains by randomizing the position of each pulse over an interval $\pm d/2$. To prevent the pulse from encroaching on its neighbors, we use "wrap-around" within its length-$d$ local domain.[6]

Thus square-wave trains can be used to represent algebraic QPs, while at the same time encoding a numerical probability, via their area. Probabilities in BNs can therefore be computed by assigning unrelated pulse trains with appropriate areas to all elementary probabilities, and then performing point-by-point multiplications and complementations in a root-to-leaf sweep that assigns pulse train representations to the marginals of all nodes. For example, a pulse train encoding of the right-hand network in Fig. 2 would begin with an assignment of unrelated pulse trains to the parameters $p, q, r, ...$, with fractional areas equal to the numerical values of these parameters. The pulse train representations of the marginals for all non-root nodes would then be computed in the following sequence (where we now think of $p, q, r, ..., P^*(B), P^*(C)$, etc., as pulse trains): $P^*(B) := p$; $P^*(C) := q$; $P^*(D) := r \cdot P^*(B) \cdot P^*(C)$; $P^*(E) := t \cdot P^*(C)$; and $P^*(F) := \mathbf{1} - (\mathbf{1} - s \cdot P^*(D)) \cdot (\mathbf{1} - u \cdot P^*(E))$. Conditioning in accord with formula (C) can then be done in general with some additional multiplications and complementations and a division. In our example, $P(B|F)$ would be computed by computing the product pulse train $P^*(B) \cdot P^*(F)$, and dividing its area by the area of the pulse train for $P^*(F)$.

The catch (to be expected for an NP-hard problem!)

---

[6]This technique was suggested by Tom Weingarten.

is that pulse train length may have to be exponential in the size of the BN to obtain a fixed accuracy. We are currently investigating anytime methods based on pulse trains, employing approximations based on successively longer pulse trains, in conjunction with techniques for gaining accuracy at reduced expense.

## 4    Discussion

We hope to have established QPs as an inherently interesting and potentially useful characterization of BN probabilities. The most noteworthy feature is the simple way in which representations of joint probabilities can be computed as weak products. Besides the computational approaches sketched above, various further techniques and potential applications based on QPs readily suggest themselves, and we conclude by mentioning some of these possibilities.

One possibility is to exploit the fact that in many Bayesian networks, many of the probabilities are small. For example, in a network of diseases and findings intended to enable diagnosis, the prior probabilities of the diseases are generally very small, and the probability of findings that are atypical for a disease (conditioned on the presence of that disease) are also small. Keeping in mind that QPs encode polynomials in the elementary probabilities, we should be able to formulate progressive approximation methods that initially neglect higher-order terms in the $*$-elimination process. We have found this quite feasible in hand-worked examples.

We are also exploring methods of "boosting" certain small probabilities as a means of gaining accuracy at low cost in certain domains, such as diagnosis. For example, suppose that we are trying to determine the posterior probability $P(D|E_1...E_n)$ for some disease with low prior probability $P(D) = p$. Then since we know that the numerator and denominator in conditioning formula (C) are both linear in $p$, we can write $P(D|E_1...E_n)$ as $(c_1 + c_2 p)/(c_3 + c_4 p)$. So theoretically if we can find values for the numerator and denominator in (C) for any two values of $p$, we can solve for $c_1, c_2, c_3, c_4$ and hence compute $P(D|E_1...E_n)$ for *any* value of $p$. Arguably, using two high values of $p$ (e.g., .5 and 1) will yield more accurate values of the constants, and hence of $P(D|E_1...E_n)$, than working directly with a small value of $p$. This method can also be extended to pairs of diseases (using 4 pairs of boosted probabilities), triples (using 8 pairs), etc.

Another natural application would be projection in probabilistic planning, i.e., determining the probability that a particular plan (or plan segment) will have certain desired effects. This lends itself naturally to a BN approach (Wellman 1990). What is particularly attractive about the use of QPs here is that they are inherently "cumulative", in a way that is well-adapted to the incremental way in which plans are built up; i.e., the QP of an anticipated effect reflects the assumptions (probabilistic "choices", in the sense of Poole (1997)) that have contributed to its derivation. As new arguments (e.g., plan segments to achieve preconditions of actions) are added, the QP of an effect can be updated (via weak products and complementations) to reflect the assumptions in these new arguments as well.

Finally, let us return briefly to the quest for a first-order probability logic in which BN-like inference is performed in a rule-based manner, rather than by explicit BN construction. In principle, the cumulativity of QPs should enable this style of inference. But can we "lift" QP calculations to quantified predicate logic? Our start would be Poole's approach (cited above), wherein the uncertainty of quantified noisy rules is modelled via independent choice variables. (Note that the elementary probabilities in an AND-OR-NOT BN could all be attached to separate root nodes, mirroring Poole's perspective.) However, we also wish to assign QPs to arbitrary quantified and logically compound formulas, something not admitted in Poole's Horn logic framework (in contrast with logics that do not rely on tacit BN-like independence assumptions – see Halpern, 2003). This seems entirely feasible. As was seen in the SAT example, assigning QPs to logical compounds is just a special case of assigning them to noisy versions of those compounds. And $\forall, \exists$-quantification can be handled by analogy with $\wedge$ and $\vee$ respectively, i.e., $P^*(\forall x \phi) = \prod_x^* P^*(\phi)$, and $P^*(\exists x \phi) = [1 - \prod_x^*(1 - P^*(\phi))]$, where the products range over assignments of domain elements to $x$.

Of course, we may ask where the $P^*(\phi)$ come from. There seem to be two cases: either $P^*(\phi)$ (for a given assignment to $x$) is determined by the QPs of other propositions that comprise it or influence it; or it is itself elementary (a choice variable). In principle we can treat any set of logically independent propositions (even complex ones) as elementary. We could use names such as $p_\phi(x)$ for them (note the dependence on the free variables of $\phi$), and use them as elements of the generalized QP algebra. Now, we may have partial knowledge about the numerical values of these $p_\phi(x)$, but a general logic may be tolerant of ignorance – even in BNs, we may only have bounds on root probabilities and conditional probabilities. So the task will be to develop ways of computing numerical bounds on QPs, where these QPs are themselves evolving dynamically as more and more knowledge is brought to bear.

But we note that (as in the SAT example) *logical truths (falsehoods) will receive QP $\simeq 1$ (0) without any*

knowledge of numerical probabilities. For example, consider the contradictory sentences $\forall x R(x)$, $\neg R(A)$. Their joint QP is $\prod_x^* P^*(R(x)) * [1 - P^*(R(A))]$, and this is easily seen to be 0 by our "resolution" rule (2). This raises the prospect of uniformly performing all reasoning, both probabilistic and deductive, by manipulation of algebraic probabilities.

**Acknowledgements**

Thanks to Tom Weingarten for preliminary quasi-probability calculations based on pulse trains, and to Jon Pakianathan and the referees for helpful pointers and comments. This work was supported in part by NSF grants IIS-0082928 and IIS-0328849.